# Self-Reasoning Assistant Learning for non-Abelian Gauge Fields Design

Jinyang Sun, [1, §] Xi Chen, [2, §] Xiumei Wang, [3] Dandan Zhu, [4, *] and Xingping Zhou [5, *]

[1] *Portland Institute, Nanjing University of Posts and Telecommunications, Nanjing 210003, China*

[2] *College of Integrated Circuit Science and Engineering, Nanjing University of Posts and Telecommunications, Nanjing 210003, China*

[3] *College of Electronic and Optical Engineering, Nanjing University of Posts and Telecommunications, Nanjing 210003, China*

[4] *Institute of AI Education, Shanghai, East China Normal University, Shanghai 200333, China*

[5] *Institute of Quantum Information and Technology, Nanjing University of Posts and Telecommunications, Nanjing 210003, China*

*§ These authors contributed equally to this work.*

*[*]Author to whom any correspondence should be addressed.*

[*]ddzhu@mail.ecnu.edu.cn

[*]zxp@njupt.edu.cn

Non-Abelian braiding has attracted substantial attention because of its pivotal role in describing the exchange behaviour of anyons, in which the input and outcome of non-Abelian braiding are connected by a unitary matrix. Implementing braiding in a classical system can assist the experimental investigation of non-Abelian physics. However, the design of non-Abelian gauge fields faces numerous challenges stemmed from the intricate interplay of group structures, Lie algebra properties, representation theory, topology, and symmetry breaking. The extreme diversity makes it a powerful tool for the study of condensed matter physics. Whereas the widely used artificial intelligence with data-driven approaches has greatly promoted the development of physics, most works are limited on the data-to-data design. Here we propose a self-reasoning assistant learning framework capable of directly generating non-Abelian gauge fields. This framework utilizes the forward diffusion process to capture and reproduce the complex patterns and details inherent in the target distribution through continuous transformation. Then the reverse diffusion process is used to make the generated data closer to the distribution of the original situation. Thus, it owns strong self-reasoning capabilities, allowing to automatically discover the feature representation and capture more subtle relationships from the dataset. Moreover, the self-reasoning eliminates the need for manual feature engineering and simplifies the process of model building. Our framework offers a disruptive paradigm shift to parse complex physical processes, automatically uncovering patterns from massive datasets.

Gauge fields play a significant role to contemporary physics, and can be classified as either Abelian or non-Abelian based on the commutativity of associated symmetry groups. Their influence has extended across various disciplines, including high-energy physics [1-4], condensed matter physics [5-7], and electromagnetism [8-10]. In recent years, non-Abelian gauge fields have emerged as a pivotal subject of inquiry, owing to their fundamental role in elucidating the exchange behavior of anyons. However, the design of such gauge fields poses formidable challenges, arising from the complex interplay of group structures, representation theory, topology, and symmetry breaking. Fortunately, circuit systems offer an optimal platform for delving into non-Abelian physics, owing to their versatile nature and ability to accommodate complex network topologies, such as non-Abelian gauge fields [11], non-Abelian inverse Anderson transitions [12], and non-Abelian topological charges and edge states [13]. These systems have been extensively utilized to investigate various topological phases and provide valuable insights into novel physical phenomena.

Deep learning as a powerful tool based on data-driven has been introduced into a number of physical research, ranging from black hole detection [14-15], gravitational lenses [16], photonic structures design [17-20], quantum many-body physics [21-23], quantum computing [24-25], topological invariants prediction [26-29], and chemical and material physics [30-32]. The applications of deep learning in physics require extensive data with or without well-defined labels, which belong to the so-called supervised learning or unsupervised learning. But most of those work is devoted to the numerical mapping (data-to-data) between input and output. The process is similar to a black box, lacking interpretability and the ability of logical reasoning. To overcome this obstacle, we set the diffusion model [33] as the reasoning module in the framework. The diffusion model can generate images by gradually denoising. The working principle is based on a diffusion process that progressively recovers image information from noise. During the training stage, the model learns how to gradually transform noise into real image data. In the generation stage, the model can start from random noise and generate images that resemble the distribution of the training data through a reverse diffusion process. Each step of the denoising process can be understood as adding noise to the image and learning how to remove these noises. It makes the operation of each step transparent and interpretable. Therefore, the reasoning module can uncover latent structures or patterns within the data, enabling operations in the latent space, such as interpolation or sample generation. This ability to discover latent structures reflects the capability to infer the intrinsic regularities of the data. Thus, the framework we proposed own strong self-reasoning capabilities.

In this work, we present a novel framework for investigating non-Abelian gauge fields within circuit systems. Our approach introduces a self-reasoning assistant learning framework, tailored to directly generate non-Abelian gauge fields. This framework demonstrates robust self-reasoning capabilities by autonomously discovering feature representations and capturing subtle relationships from raw data. Moreover, it showcases end-to-end learning prowess, obviating the need for manual feature engineering and streamlining the model-building process. Specifically, we introduce a circuit-based platform for showcasing the relevant characteristics of non-Abelian phenomena. Our approach directly generates complex non-reciprocal circuit diagrams on-demand. These diagrams include circuits with 4 $C_3$ units and 3 coupled connections arranged in a chain configuration. The core advantage of this method lies in its ability to directly generate circuit design diagrams based on input design structure information, significantly reducing steps of circuit design. To ensure the reliability of our findings, we select 500 pairs of network predictions and corresponding theoretical results for comparison. Besides, we conduct experiments by translating partial predicted circuit diagrams onto Printed Circuit Board (PCB). Through theoretical and experimental validation, our network achieves an accuracy rate of 77.4%, providing an intuitive and efficient solution for non-Abelian gauge field circuit design.

We emphasize the proposition of generating expected circuit structures through a multimodal network, hence we do not need to prepare an extremely large dataset. Nevertheless, we still have produced more than 4,000 sets of circuit diagrams and structure information for dataset construction and model training. The results indicate that our model can learn effective patterns from relatively small datasets and generate non-Abelian circuit structures on-demand. In practical applications, our approach offers a feasible solution, particularly in situations where computing resources or time are limited.

## Result

### Design of Non-Abelian Gauge Field Circuit Structure

In contrast to the commutative properties inherent in Abelian theories, non-Abelian theories exhibit captivating behaviors due to their non-commutative nature of operations. Specifically, we start with supposing there are two identical particles in a two-dimensional space. Then the wave function can undergo arbitrary phase changes when one particle exchanges with the other particle in a counterclockwise manner:

$$\psi(r_1, r_2) \longrightarrow e^{i\theta}\psi(r_1, r_2), \tag{1}$$

where $\psi(r_1, r_2)$ is the original wave function, and $\theta$ is the phase shift. As the second counterclockwise exchange may not necessarily return to the initial state, it will result in a non-trivial variation phase, as shown below:

$$\psi(r_1, r_2) \longrightarrow e^{2i\theta}\psi(r_1, r_2), \tag{2}$$

where $\theta = 0, \pi$ correspond to bosons and fermions, respectively. Particles possessing alternative 'statistical angles' are referred to as anyons [34]. We refer to such particles as anyons with statistics $\theta$.

When turning our attention to the general case of *N* particles, a more intricate braid-like structure emerges. The trajectories that transition these particles from their original positions $R_1, R_2, ..., R_N$ at time $t_i$ to their designated positions $R_1, R_2, ..., R_N$ at time $t_f$ are intricately linked to the components of the braid group $\beta_N$ . An element of the braid group can be visualized by imagining particle trajectories as world lines (or strings) in 2+1 dimensional spacetime, where these trajectories initiate from specific starting points and culminate at defined endpoints. The time direction is represented vertically, with the initial time at the bottom and the final time at the top, as depicted in Fig. 1(a-c). Through smooth deformations, an element of the *N*-particle braid group is defined as an equivalence class comprising such trajectories, where each trajectory represents a distinct path in the spacetime framework. To represent an element of the class, it is necessary to order the initial and final points along the lines of initial and final times. In drawing the trajectories, care must be taken to distinguish between one line crossing over another, which corresponds to clockwise or counterclockwise exchanges. At any intermediate time slice, there must be intersections with *N* lines, and these lines cannot 'turn back' to prevent the implied creation or annihilation of particles at intermediate stages. The multiplication of two elements in the braid group is simply the successive execution, and it is evident from the diagram that the order of multiplication matters, as the group is non-abelian, implying that multiplication is non-commutative.

Braiding statistics are linked with high-dimensional representations, which will emerge in the set of degenerate states *g* when particles are fixed at positions. An orthogonal basis $\psi_a$ with a = *1, 2, …, g* is defined for these degenerate states. The elements of the braid group include $\sigma_1$ and $\sigma_2$. The $\sigma_1$ exchanges particles 1 and 2, while the $\sigma_2$ exchanges particles 2 and 3. The actions are represented by $g \times g$ unitary matrices $\rho(\sigma_1)$ and $\rho(\sigma_2)$:

$$\psi_a \to [\rho(\sigma_1)]_{\alpha\beta}\psi_\beta \, , \; \psi_a \to [\rho(\sigma_2)]_{\alpha\beta}\psi_\beta \, , \tag{3}$$

where both $[\rho(\sigma_1)]_{\alpha\beta}$ and $[\rho(\sigma_2)]_{\alpha\beta}$ are $g \times g$ unitary matrices, defining unitary transformations within the subspace of degenerate ground states. If $\rho(\sigma_1)$ and $\rho(\sigma_2)$ do not commute, i.e., $[\rho(\sigma_1)]_{\alpha\beta}[\rho(\sigma_2)]_{\beta\gamma} \neq [\rho(\sigma_2)]_{\alpha\beta}[\rho(\sigma_1)]_{\beta\gamma}$, the particles obey non-Abelian braid statistics. The braid statistics of particles remain Abelian state unless they commute for every particle exchange. Braided quasi-particles induce non-trivial rotations within the degenerate multi-quasi-particle Hilbert space, when they do not commute.

In systems possessing arbitrary sub particles, it is often necessary to consider multiple types of anyons. Fusion refers to the process of combining two anyons into a single particle entity. In the Abelian scenario, the fusion rules are straightforward [35]. For instance, $2\pi n/m \times 2\pi k/m = 2\pi(n+k)/m$, where the operation $a \times b$ denotes the fusion of *a* and *b*. However, there may not be a unique method to combine topological quantum numbers for non-Abelian anyons. For example, two particles each with a spin of 1/2 could be combined to form a particle with either spin 0 or spin 1. These different possibilities are known as distinct fusion channels, which is typically represented by [34]:

$$\varphi_a \times \varphi_b = \sum_c N_{ab}^c \varphi_c \, , \tag{4}$$

where both $\varphi_a$ and $\varphi_b$ denote the fusion channels, and $N_{ab}^c$ represents the multiplicity of fusion. Such terminology clarifies that in non-Abelian anyon scenarios, fusion between particles of type a and type b might result in a particle of type *c* ($N_{ab}^c \neq 0$).

In our work, we focus on exploring innovative approaches for developing complex and efficient circuit, guided by non-Abelian theoretical principles. This exploration occurs within the realm of circuit design, where understanding non-commutative theories is paramount. As in the field of circuit design, the principles of non-commutative theory are crucial. The methods of coupling and the employed sequences play a pivotal role in shaping the behavior and functionality of non-Abelian circuits. And the choice of couplings, along with the sequence in which these couplings are applied, determines the emergent properties of the circuits as well.

Our approach involves utilizing a circuit coupling module that implements non-Abelian tunneling in the form of Pauli matrices. We have explored a strategy that constructs a doubly degenerate space using basic electronic components and vector potentials of the Pauli matrix form. As shown in Fig. 1(e), we consider a configuration where three identical components (either capacitors or inductors) are connected at the heads in cell-*m* and cell-*n* to form a triangle, which exhibits $C_3$ rotational symmetry [11, 36]. For the $C_3$ unit, it possesses a two-dimensional irreducible representation, whose basis functions are complex conjugates of each other. Consequently, the circuit's doubly degenerate eigenstates can be selected as the basis for the pseudospin space. Our design commences with a chain circuit comprising four $C_3$ units, with detailed derivations provided in Appendix A.

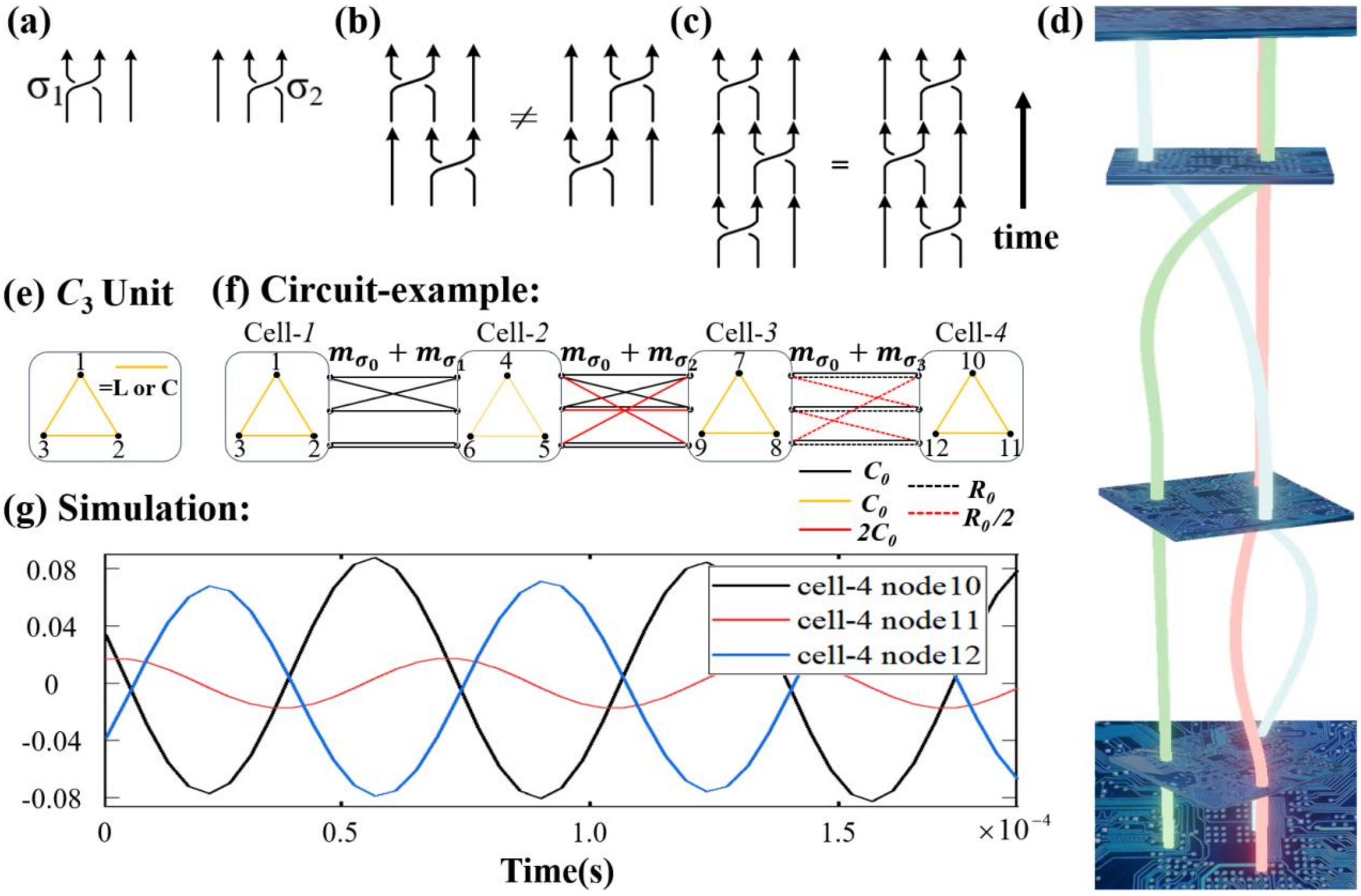

Fig. 1 Construction and design of non-Abelian gauge fields in circuits. (a) The two elementary braid operations $\sigma_1$ and $\sigma_2$ on three particles. (b) Non-Abelian braiding theory, where the braiding relation is $\sigma_2\sigma_1\neq\sigma_1\sigma_2$. (c) Abelian braiding theory, with the general braiding relation being $\sigma_i\sigma_{i+1}\sigma_i=\sigma_{i+1}\sigma_i\sigma_{i+1}$. (d) Three-dimensional non-Abelian braiding diagram. (e) Three components (capacitors or inductors) with the same parameters are connected head to tail to form a triangle. The nodes are denoted as *1*, *2* and *3* (f) An example of a non-reciprocal circuit generating non-Abelian phase factors. The capacitor $C$ = 2.7 nF with a tolerance of ±20%, and resistor $R$ = 1 kΩ with a tolerance of ±1%. In each cell, nodes are grounded through capacitors or resistors, ensuring the same resonant frequency across all four units. (g) Simulation results for the output voltages at the three nodes of cell-*4*.

Subsequently, we expand the original circuit to include four $C_3$ modules, thereby enhancing the diversity and complexity of the coupling. As depicted in Fig. 1(f), we utilize an inductive circuit as an example to conduct a non-Abelian theoretical analysis. Given the input current is applied at cell-*1* and the output voltage is measured at cell-*4*, it is practical to compute the current-voltage transfer function between cell-*1* and cell-*4*. Upon calculation, the transfer function is as follows:

$$\tilde{v}_{out} = Z_{out}\tilde{i}_{in} = (t_1 t_2 t_3 / \gamma \oplus h_{21}h_{32}h_{43} / \Delta)\tilde{i}_{in}\,, \tag{5}$$

where $\psi = \lambda_1^2(-\lambda_1^2 + t_1^2 + t_2^2 + t_3^2) - t_1^2 t_3^2$ , $\Delta = \lambda_2^2(h_{12}h_{21} + h_{23}h_{32} + h_{34}h_{43} - \lambda_2^2) - h_{12}h_{21}h_{34}h_{43}$ , $\tilde{v}_{out}$ represents the output voltage at cell-*4*, $\tilde{i}_{in}$ denotes the input current at cell-*1*. The Alternating currents $\tilde{i}_{in} = (\tilde{i}_1, \tilde{i}_2, \tilde{i}_3) = i_0\varphi_0 + i_s(\cos\frac{\eta}{2}\varphi_{S_1} + \sin\frac{\eta}{2}e^{i\kappa}\varphi_{S_2})$ , where $\varphi_0 = (1,1,1)^T / \sqrt{3}$ , $\varphi_{S_1} = (\varepsilon, \varepsilon^*, 1)^T / \sqrt{3}$, $\varphi_{S_2} = (\varepsilon^*, \varepsilon, 1)^T / \sqrt{3}$, η and κ denote the positioning of the spin vector on the Bloch sphere, which carries the information of spin. We take $i_0$=0, $i_s$=0.02, η=π/3 and κ=π/5, resulting in $(\tilde{i}_1, \tilde{i}_2, \tilde{i}_3) = (0.005\measuredangle 147^\circ, 0.012\measuredangle -149^\circ, 0.015\measuredangle 13^\circ)$ for cell-*1*, with each current having a frequency of 15 kHz. More detailed derivations are provided in Appendix B.

**Self-Reasoning Assistant Learning Framework**

To facilitate the construction of non-Abelian circuits and the computation of their transfer functions, we have developed a series of non-Abelian circuit models that include four $C_3$ structures and three coupling modules. During data collection, it is stipulated that all coupling connections must be selected from Appendix C. Theoretically, the total number of circuit combinations is approximately $10^4$ (calculated as the product of 48×47×46), with each coupling method occurring only once. To reduce computational complexity, we select 4,096 sets of circuit diagrams and structure information for dataset construction and model training. Our training dataset consists of circuits along with their transfer function data, including both the images and the structure information. The images show schematic diagrams of circuits with varying component values and coupling modes, while the accompanying structure information provide details on the transfer functions, input currents, and output voltages of these circuits.

Subsequently, an appropriate network model is proposed for the circuit data. To address the complexity of the training dataset and meet on-demand design requirements, we introduce a text-to-image diffusion model that capitalizes on the large Transformer language model's text comprehension abilities [37]. Distinguishing itself from other language models, it allocates the training focus on the text encoder rather than on the image generation aspect, which equips itself with the ability for self-reasoning assistant learning. This feature is particularly beneficial for non-Abelian gauge fields design, facilitating the direct derivation of circuit design diagrams from input structure information. due to the extensive parameterization of T5, the framework can yield satisfactory outcomes even without fine-tuning.

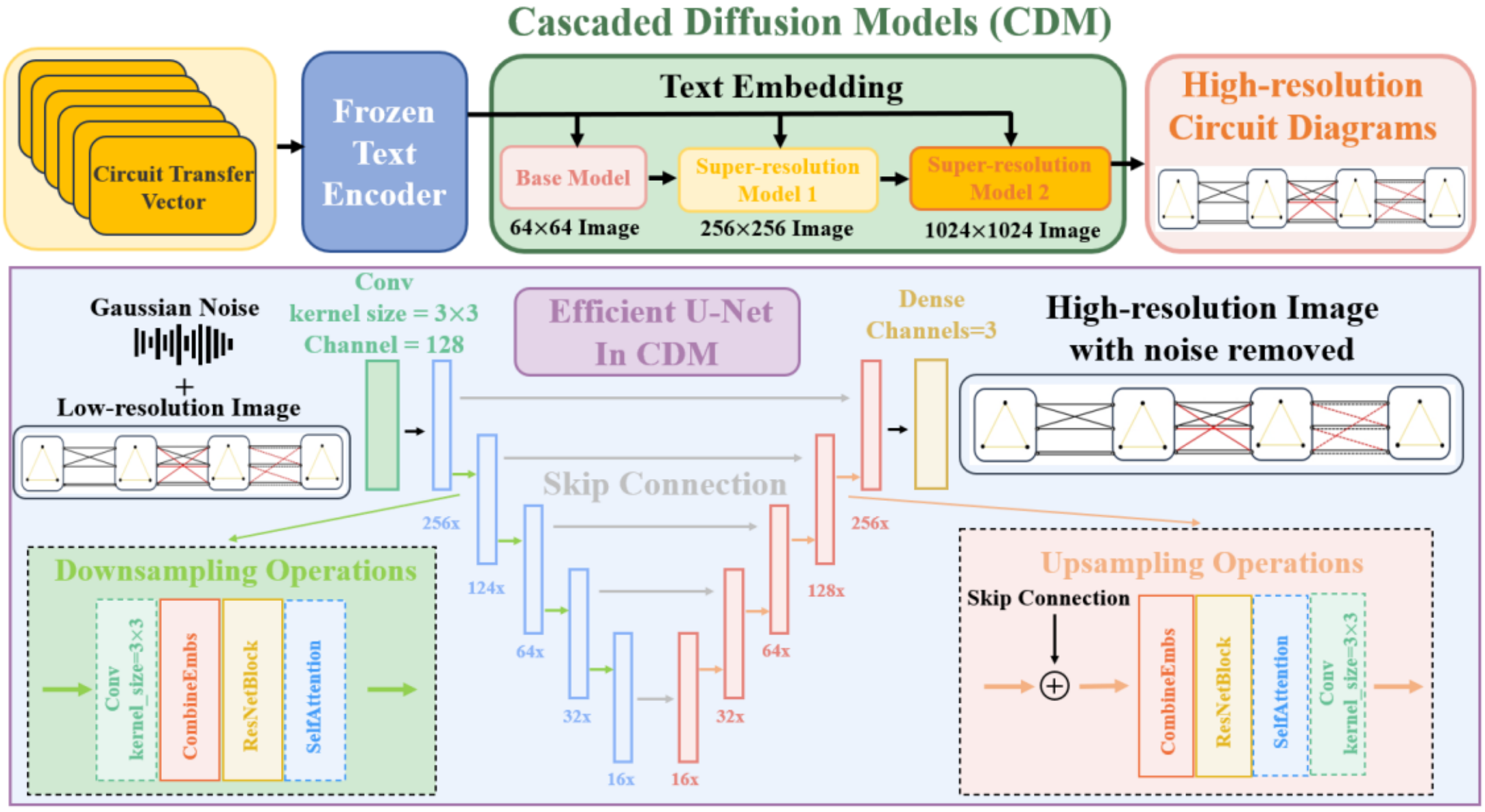

Fig. 2 Illustrative procedure. The cascaded diffusion model of primarily consists of a Base model and two Super-resolution models, each Super-resolution model being constituted by an efficient U-Net. The purple rectangle provides a detailed illustration of the efficient U-net process used in the Cascade Diffusion Models, as shown in the green box above.

Our self-reasoning framework is powered by its advanced Text Encoding, which uses Transformer models to interpret and operationalize text-based inputs. This capability enables the model to autonomously generate detailed circuit designs from descriptions, adapting to diverse and complex scenarios. The integration of a large-scale language model also enhances continuous learning and streamline the design process without extensive manual intervention. As shown in Fig. 2, our framework includes two primary sections: the Text Encoding segment and the Cascaded Diffusion Model. In the Text Encoder segment, we utilize a fine-tuned T5-large model (with a parameter count of 11 billion) [38], enabling the model to achieve a higher level of comprehension of the text we provide. The Cascaded Diffusion Model is further divided into two parts: the Base model and the Super-resolution model. The essence of the Base model is a typical 64 × 64 U-Net structure, which can iteratively convert Gaussian noise into samples from the learned data distribution through a denoising process. Since Diffusion Models are latent variable models, their latent variable $z=\{z_t \mid t\in[0,1]\}$ adheres to a forward process $q(z \mid x)$ that starts from the data $x\sim p(x)$. Consequently, the forward process is a Gaussian process that satisfies Markovian structure [37]:

$$q(z_t \mid x)=N(z_t;\alpha_t x,\sigma_t^2 I),\ q(z_t \mid z_s)=N(z_t;(\alpha_t/\alpha_s)z_s,\sigma_{t|s}^2 I), \tag{6}$$

where $0\le s<t\le 1$, $\sigma_{t|s}^2=(1-e^{\lambda_t-\lambda_s})\sigma_t^2$, $\alpha_t$ and $\alpha_s$ represent noise that follows a Gaussian distribution. For instance, $\alpha_t$ and $\sigma_t$ specify a differentiable noise schedule, whose log signal-to-noise ratio $\lambda_t=\log[\alpha_t^2/\sigma_t^2]$ decreases over time *t*, until $q(z_1)\approx N(0,I)$, where *N(0, I)* denotes the Gaussian distribution with a mean of 0 and a variance of *I*. Another prominent feature of the Base model is the Classifier-Free Guidance, which is the substitution of an explicit classifier with an implicit one, thereby obviating the need for direct computation of the explicit classifier and its gradients. The classifier operates by randomly dropping the condition during the training process, allowing for both conditional and unconditional sampling inputs. Both types of inputs are fed into the same diffusion model, thus enabling it to possess the capability to generate both conditionally and unconditionally. And the final sampling process can be articulated as follows [37]:

$$\tilde{\varepsilon}_\theta(z_t,c)=\omega\varepsilon_\theta(z_t,c)+(1-\omega)\varepsilon_\theta(z_t), \tag{7}$$

where $\varepsilon_\theta(z_t,c)$ and $\varepsilon_\theta(z_t)$ respectively represent the conditional and unconditional predictions, $\omega$ is the guidance weight. For the Super-resolution model, we initially process the 64 × 64 images generated by the Base model through an efficient U-Net to upscale them to a size of 256 × 256, as detailed in Fig. 2. Subsequently, another efficient U-Net receives the complete 256 × 256 image as low-resolution input and returns an up sampled 1024 × 1024 image as the final output. This approach directly enhances efficiency while, through the augmentation of noise, it improves the model's robustness in controlling distortions. After ample training, our network ultimately acquires the capability to stably generate circuit diagrams based on input structure information, showcasing its proficiency in self-reasoning capabilities.

To verify the reliability and accuracy of the model, we input 500 datasets and have the model generate circuit diagrams that meet the requirements. Subsequent theoretical calculations and statistical validation show that 387 of the generated circuit diagrams are entirely correct, 79 diagrams have minor errors with fewer than five incorrect connections, and only 34 diagrams exhibit more significant errors. We opt for this direct evaluation approach because circuit design diagrams are either correct or incorrect, with no gray area. Therefore, comparing predicted diagrams directly with theoretical ones is the most suitable method to assess accuracy. The absolute correctness rate has reached 77.4%, which indicates the favorable training outcomes and the effectiveness of our network model. Partial prediction results are illustrated in Fig. 3(a).

## Experimental Validation

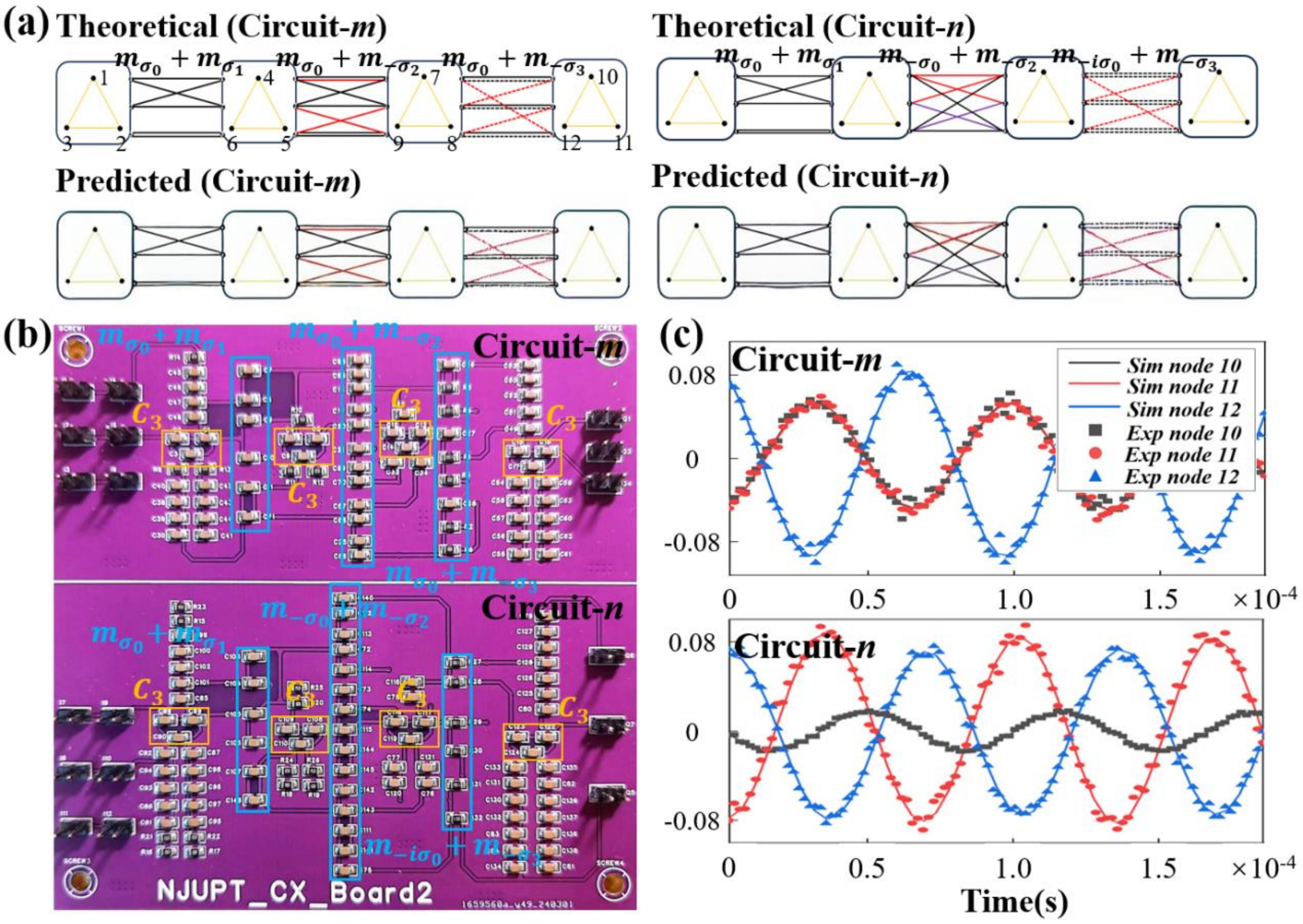

Fig. 3 Generate non-Abelian circuits based on specific features. (a) We provide the model with characteristics of circuits-*m* and -*n*, including transfer function coefficient vectors, node connection components, and circuit composition to generate predicted circuit diagrams. (b) The constructed circuits-*m* and -*n* on a PCB. (c) The output voltages at three nodes of cell-*4* are measured with the same input current to cell-*1*, where the dots represent experimental data and the solid lines represent theoretical results. The capacitor $C_0$ is rated at 2.7 nF with a tolerance of ±20%, and the resistor $R_0$ is 1 kΩ with a tolerance of ±1%.

Initially, we randomly supply circuit features to the model, including the transfer function coefficient vector, node connection components, and circuit configuration. Based on these characteristics, the model generates predicted circuit diagrams, as exemplified in Fig. 3(a). We select two representative examples from the structure information inputs used to generate circuit-*m* and circuit-*n*. For circuit-*m*, the inputs are: (1) transfer function coefficient vector [8.57e-18 + 7.06e-17i, 0.05 - 0.13i, 0.10 - 0.01i], and (2) node connection components 5C + 1.5R. For circuit-*n*, the inputs are: (1) transfer function coefficient vector [-3.44e-18 - 1.82e-17i, 0.06 + 0.02i, 0.06 + 0.02i], and (2) node connection components 5C + 2.5R. We construct circuit- m and -n on PCB boards for experimental verification, as depicted in Fig. 3(b). Besides, a constant input current $(\tilde{i}_1,\tilde{i}_2,\tilde{i}_3)=(0.005\measuredangle 147^{\circ},0.012\measuredangle -149^{\circ},0.015\measuredangle 13^{\circ})$ is maintained at cell-*1*, and the output voltages of cell-*4* are measured. As illustrated in Fig. 3(c), the dots represent experimental data and lines indicate simulation results, which prove that the network-predicted circuits match the theoretical circuits, effectively meeting the design requirements.

Subsequently, we examine three circuits comprised of cells-*1*, -*2*, -*3*, -*4* and two connecting modules per unit, to explicitly characterize the non-reciprocal nature of non-Abelian gauge fields, shown in Fig. 4(a). It demonstrates the phase tuning of signals in real space in a non-Abelian manner. As the signal passes through the circuit, its pseudospin components are confined within this degenerate pseudospin space and acquire phase modulation in the form of a non-Abelian gauge field. For the same initial state $\tilde{i}_0$, passing through circuit-*1*, -*2*, and -*3*, different final states $\tilde{v}_{10}$, $\tilde{v}_{11}$ and $\tilde{v}_{12}$ can be obtained, as shown in Fig. 4(b). The detailed calculation processes of the transfer functions for circuits *1*, *2*, and *3* are presented in Appendix D.

We input the descriptive structure information corresponding to circuits *1*, *2*, and *3* into the model to predict the circuit models, with the predictions displayed in Fig. 4(c). To validate the predictions, we construct the predicted circuits *1*, *2*, and *3* on a PCB, as illustrated in Fig. 4(d). Alternating current is applied at three nodes of cell-*1*, and the output voltage is measured at the node of cell-*4*. We compare the predicted circuit diagrams with simulated theoretical ones, as depicted in Fig. 4(b). The result indicates a good match between the experimental and theoretical curves, with the DC component removed from the experimental results. The capacitor $C_0$ is 2.7 nF with a tolerance of ±20%. The resistor $R_0$ is 1 kΩ with a tolerance of ±1%. The current frequency is set at 15 kHz, generated by an OPA549 module. The output voltage is measured with a SIGLENT SDS2202X Plus oscilloscope. A detailed list of the experimental instruments is provided in Appendix E.

To compare with theoretical results, we exclude the DC components from the experimental data. The generation of DC errors mainly stems from that we provide current to three input ports in steps using a single current source. Due to the linear nature of the circuit, the total output voltage is the sum of the voltages measured in these three steps. However, this measurement method comes at the cost of relatively large cumulative errors. The DC component measurement will decrease when three current sources input simultaneously. Since the critical information in the experimental data involves the amplitude and phase of the AC components, the presence of DC components does not affect our conclusions. The experimental results indicate the successful training of our network model tailored for non-Abelian circuits, with its effectiveness and accuracy verified through simulation and experimentation.

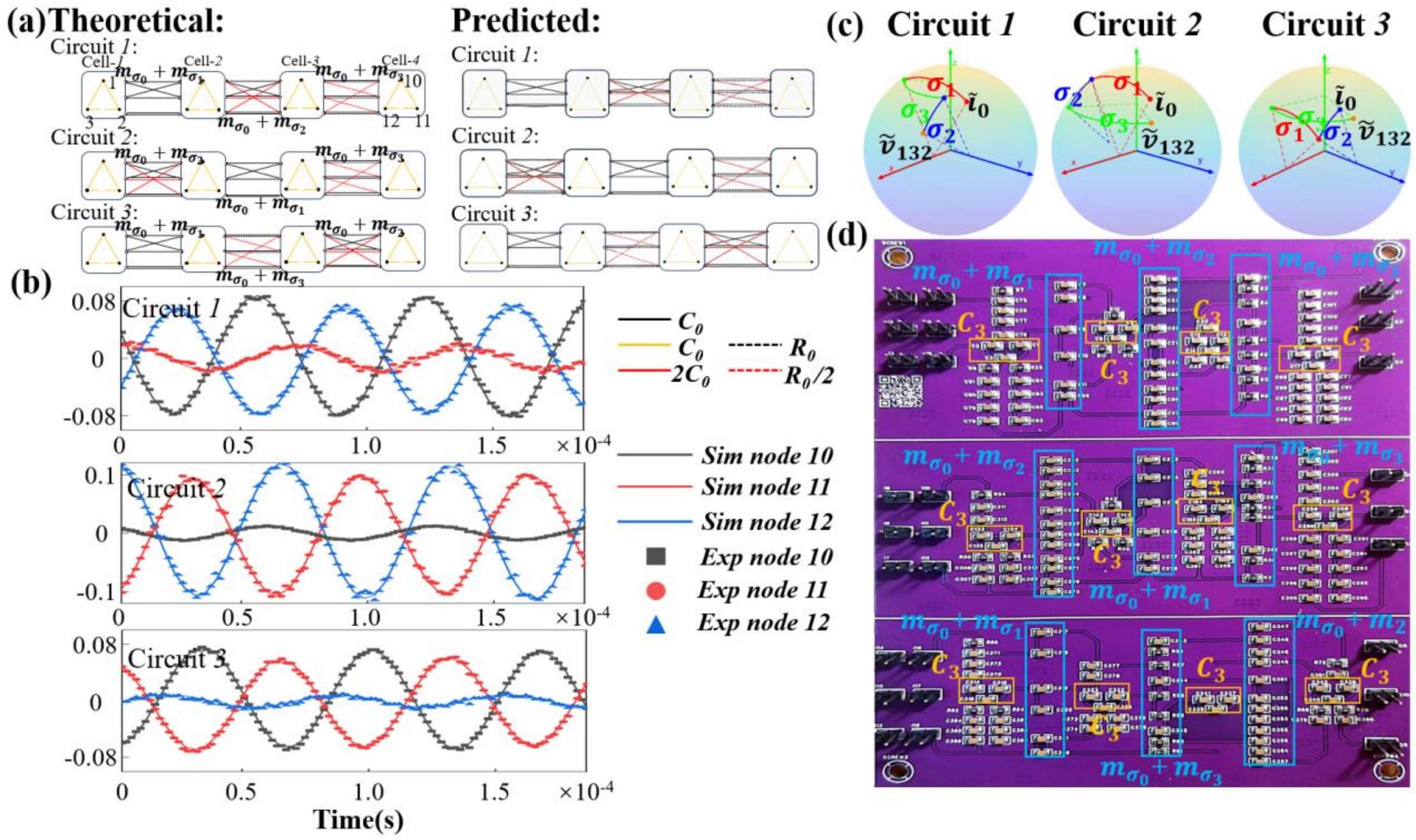

Fig. 4 Generate non-reciprocal circuits with non-Abelian phase factors as required. (a) Schematic diagrams for circuits-*1*, -2, and -3, showing capacitors arranged in triangular configurations within units 1, 2, 3, and 4. Each unit is grounded through capacitors and resistors to maintain identical on-site frequencies. Coupling modules composed of $m_{\sigma_0}+m_{\sigma_1}$, $m_{\sigma_0}+m_{\sigma_2}$, and $m_{\sigma_0}+m_{\sigma_3}$ are swapped among the four $C_3$ units to form Circuit *1*, *2*, and *3*. (b) Output voltages at three nodes of cell-*4* are measured under a constant input current to cell-*1*, with the dots indicating experimental data and the lines theoretical results. (c) Diagram illustrating different voltage end states under a non-Abelian gauge field, initiated by the same current. (d) Circuits-*1*, -*2*, and -*3* are constructed on a PCB.

## Discussion

The presented framework constructs a low-dimensional latent space of non-Abelian gauge circuits, by self-reasoning assistant learning technologies. It is powered by an advanced Text Encoding using Transformer models to interpret and operationalize text-based inputs, which can iteratively convert Gaussian noise into samples from the learned data distribution through a denoising process. This process enables the framework to autonomously generate detailed circuit designs from descriptions, adapting to diverse and complex scenarios. Our framework can be adopted in the data-dependent scenarios which are difficult to describe with formulas and extended to other physical fields such as optics, acoustics, mechanics. With the advancement in the interpretability of deep learning methods in relation to physics, we can leverage the tool to uncover deeper laws of the physical world.

**Acknowledgements**

The authors thank for the support by National Natural Science Foundation of China under (Grant 62001289), NUPTSF (Grants No. NY220119, NY221055). We thank Professor Xiaofei Li for useful discussions.

## APPENDIX A: DERIVATIONS OF THE TRANSMISSION MATRIX FOR A CHAIN CIRCUIT OF THREE $C_3$ UNITS

We take one circuit as an example, which is shown below:

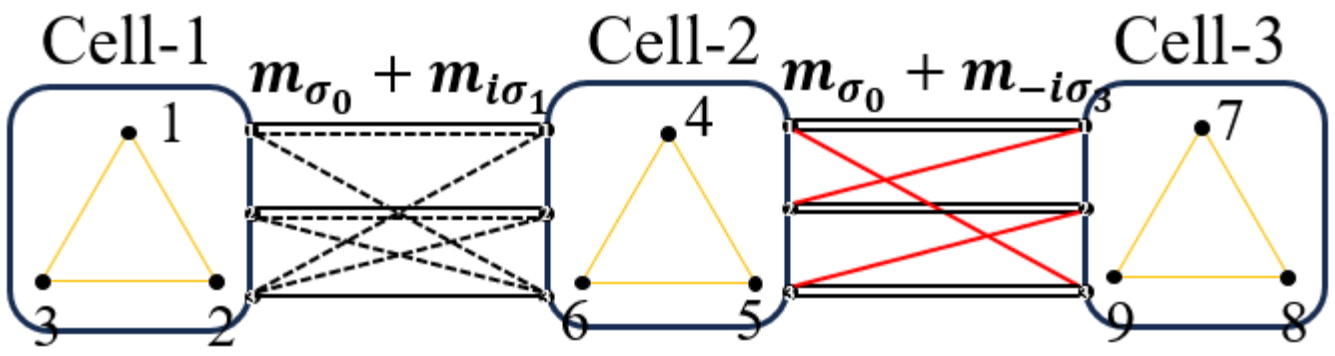

FIG. S1 A schematic diagram of a circuit. Capacitors in units *1*, *2*, and *3* form a triangular configuration. Nodes in units 1 and 3 are grounded through capacitors and resistors to ensure all three units maintain the same on-site frequency. The coupling modules are connected by $m_{\sigma_0} + m_{i\sigma_1}$ and $m_{\sigma_0} + m_{-i\sigma_3}$ to form a chain circuit.

We take one circuit as an example, shown in Fig. S1. The Kirchhoff's equations for the circuit are as follows:

$$\boldsymbol{I} = Y\boldsymbol{V} = \begin{pmatrix} Y_0 & Y_{12} & 0 \\ Y_{12}^T & Y_0 & Y_{23} \\ 0 & Y_{23}^T & Y_0 \end{pmatrix} \boldsymbol{V}, \tag{A1}$$

where $\boldsymbol{V} = (v_1, v_2, v_3, ..., v_7, v_8, v_9)$, $\boldsymbol{I} = (i_1, i_2, i_3, ..., i_7, i_8, i_9)$,

$$Y_o \,|= y_{C_0} M_0 - (y_{C_{11}} + 2y_{R_{11}} + y_{C_{31}} + 3y_{C_{32}}),$$

$$Y_{12} \,|= y_{C_{11}} M_{\sigma_0} + y_{R_{11}} M_{i\sigma_1},$$

$$Y_{23} \,|= y_{C_{31}} M_{\sigma_0} + y_{C_{32}} M_{-i\sigma_3}.$$

Subsequently, we make a transformation of the above formula by using the *U* matrix to obtain

$$\begin{pmatrix} \tilde{i}_{1,2,3} \\ \tilde{i}_{4,5,6} \\ \tilde{i}_{7,8,9} \end{pmatrix} = \begin{pmatrix} U^\dagger Y_0 U & U^\dagger Y_{12} U & 0 \\ U^\dagger Y_{12}^T U & U^\dagger Y_0 U & U^\dagger Y_{23} U \\ 0 & U^\dagger Y_{23}^T U & U^\dagger Y_0 U \end{pmatrix} \begin{pmatrix} \tilde{v}_{1,2,3} \\ \tilde{v}_{4,5,6} \\ \tilde{v}_{7,8,9} \end{pmatrix}, \tag{A2}$$

where characteristic matrix $U = \frac{1}{\sqrt{3}} \begin{pmatrix} 1 & \varepsilon & \varepsilon^* \\ 1 & \varepsilon^* & \varepsilon \\ 1 & 1 & 1 \end{pmatrix}$, $\varepsilon = e^{i2\pi/3}$, $U^\dagger$ is the conjugate transpose matrix of characteristic matrix *U*. Simultaneously, the current and voltage denoted with a tilde(~) at the top are calculated based on the characteristic functions of the $C_3$ symmetry group as the basis vectors, whereas the current and voltage without a tilde(~) represent the actual current and voltage at the nodes of the circuit.

Here, we focus primarily on the right side of Eq. (A2), where the basis vectors have now transformed into $(\tilde{v}_{1,2,3}, \tilde{v}_{4,5,6}, \tilde{v}_{7,8,9})^T$ , with $\tilde{v} = U^{\dagger} v$ . Taking $U^{\dagger} Y_{12} U$ as an example, it is defined as:

$$
\begin{aligned}
Y_{T12} = U^{\dagger} Y_{12} U &= \begin{pmatrix} y_{C_{11}} + 2y_{R_{11}} & 0 & 0 \\ 0 & y_{C_{11}} & -y_{R_{11}} \\ 0 & -y_{R_{11}} & y_{C_{11}} \end{pmatrix} \\
&= (y_{C_{11}} + 2y_{R_{11}}) \oplus \begin{pmatrix} y_{C_{11}} & -y_{R_{11}} \\ -y_{R_{11}} & y_{C_{11}} \end{pmatrix},
\end{aligned}
\tag{A3}
$$

where $y_{C_0}$ represents the admittance of C0, $\oplus$ denotes the direct sum. The term preceding $\oplus$ is referred to as the constant representation space, and the subsequent term is termed the pseudo-spin space. Applying the same operation, we can derive $Y_{T0}$ and $Y_{T23}$ . The operation transforms Eq. (A2) into a pseudo-spin space and reorders the bases, segregating the constant representation space from the pseudo-spin space, which is detailed below:

$$
Y = \begin{pmatrix} \lambda_1 & t_1 & 0 & 0 \\ t_1 & \lambda_1 & t_2 & 0 \\ 0 & t_2 & \lambda_1 & t_3 \\ 0 & 0 & t_3 & \lambda_1 \end{pmatrix} \oplus \begin{pmatrix} \lambda_2 & h_{12} & 0 & 0 \\ h_{21} & \lambda_2 & h_{23} & 0 \\ 0 & h_{32} & \lambda_2 & h_{34} \\ 0 & 0 & h_{43} & \lambda_2 \end{pmatrix}, \tag{A4}
$$

where

$$
\begin{gathered}
\lambda_{1S} |= -s(C_{11} + C_{31} + 3C_{32}) - 2/R_{11} \\
\lambda_{2S} |= \lambda_{1S} - 3sC_0 \\
t_1 |= sC_{11} + 2/R_{11} \\
t_2 |= s(C_{31} + 3C_{32}) \\
t_3 |= sC_0 + 3/R_0 \\
h_{12} |= sC_{11}\sigma_0 - 1/R_{11}\sigma_1 \\
h_{21} |= h_{12} \\
h_{23} |= s(C_{31}\sigma_0 - \sqrt{3}C_{32} i\sigma_3) \\
h_{32} |= s(C_{31}\sigma_0 + \sqrt{3}C_{32} i\sigma_3)
\end{gathered}
$$

Continuing the derivation, we examine the relationship between the input current and the output voltage. It is known from $\boldsymbol{I} = Y\boldsymbol{V}$ that

$$
\boldsymbol{V} = Z\boldsymbol{I},\ Z = Y^{-1}, \tag{A5}
$$

$$Z = Y^{-1} = \begin{pmatrix} \lambda_1 & t_1 & 0 \\ t_1 & \lambda_1 & t_2 \\ 0 & t_2 & \lambda_1 \end{pmatrix}^{-1} \oplus \begin{pmatrix} \lambda_2 & h_{12} & 0 \\ h_{21} & \lambda_2 & h_{23} \\ 0 & h_{32} & \lambda_2 \end{pmatrix}^{-1}, \tag{A6}$$

And we take the representation of space with constants as an example,

$$\begin{pmatrix} \lambda_1 & t_1 & 0 \\ t_1 & \lambda_1 & t_2 \\ 0 & t_2 & \lambda_1 \end{pmatrix}^{-1} = \begin{pmatrix} -\frac{\lambda_1^2 - t_2^2}{\lambda_1\sigma_1} & \frac{t_1}{\sigma_1} & -\frac{t_1 t_2}{\lambda_1\sigma_1} \\ \frac{t_1}{\sigma_1} & -\frac{\lambda_1}{\sigma_1} & \frac{t_2}{\sigma_1} \\ -\frac{t_1 t_2}{\lambda_1\sigma_1} & \frac{t_2}{\sigma_1} & -\frac{\lambda_1^2 - t_1^2}{\lambda_1\sigma_1} \end{pmatrix}, \tag{A7}$$

$$(\tilde{v}_1, \tilde{v}_4, \tilde{v}_7) = \begin{pmatrix} -\frac{\lambda_1^2 - t_2^2}{\lambda_1\sigma_1} & \frac{t_1}{\sigma_1} & -\frac{t_1 t_2}{\lambda_1\sigma_1} \\ \frac{t_1}{\sigma_1} & -\frac{\lambda_1}{\sigma_1} & \frac{t_2}{\sigma_1} \\ -\frac{t_1 t_2}{\lambda_1\sigma_1} & \frac{t_2}{\sigma_1} & -\frac{\lambda_1^2 - t_1^2}{\lambda_1\sigma_1} \end{pmatrix} (\tilde{i}_1, \tilde{i}_4, \tilde{i}_7)^T. \tag{A8}$$

Here, we focus exclusively on the output voltage $\tilde{v}_7$, given that only input $\tilde{i}_1$ is active, denoted as $\tilde{i}_1 \neq 0$, $\tilde{i}_4 \neq 0$ and $\tilde{i}_7 \neq 0$. Therefore, $\tilde{v}_7 = -\frac{t_1 t_2}{\lambda_1\sigma_1}\tilde{i}_1$, corresponding to the constant representation space of $Z_{out}$ in Eq. (A6). By employing the same method, the relationship between $\tilde{v}_8$ and $\tilde{v}_9$ with respect to $\tilde{i}_2$ and $\tilde{i}_3$ in the pseudo-spin space can be determined. It is discovered that starting from Kirchhoff's equation $\boldsymbol{I} = \boldsymbol{YV} = \begin{pmatrix} Y_0 & Y_{12} & 0 \\ Y_{12}^T & Y_0 & Y_{23} \\ 0 & Y_{23}^T & Y_0 \end{pmatrix} \boldsymbol{V}$ and deriving based on $\tilde{i}_{4,5,6,7,8,9} = 0$, identical results can be obtained as well.

Given the input current is applied at cell-*1* and the output voltage is measured at cell-*3*, it is practical to compute the current-voltage transfer function between cell-*1* and cell-*3*. Upon calculation, the transfer function is as follows.

$$\tilde{v}_{out} = Z_{out}\tilde{i}_{in} = (\lambda_{1s}^{-1}/2 \oplus (-h_{32}\xi h_{21}))\tilde{i}_{in}, \tag{A9}$$

where $\xi = (\lambda_{2S}h_{23}h_{32} + \lambda_{2S}h_{21}h_{12} - \lambda_{2S}^3)^{-1}$, $\tilde{V}_{out}$ represents the output voltage at cell-*4*, $\tilde{i}_{in}$ denotes the input current at cell-*1*.

## APPENDIX B: DETAILS OF THE FIG. 1(F) CIRCUIT EXAMPLE

The Kirchhoff current equations for the circuit in Fig. 1(e) are

$$\begin{pmatrix} i_1 \\ i_2 \\ i_3 \end{pmatrix} = \frac{1}{i\omega L}\begin{pmatrix} -2 & 1 & 1 \\ 1 & -2 & 1 \\ 1 & 1 & -2 \end{pmatrix}\begin{pmatrix} v_1 \\ v_2 \\ v_3 \end{pmatrix}, \tag{A10}$$

where $v_{1,2,3}$ represent the voltages at the nodes and $i_{1,2,3}$ represent the currents flowing into the nodes. $L$ denotes the inductance, and $\omega$ is the frequency of the alternating current signal. The 3 × 3 matrix on the right-hand side of Eq. (A10) is abbreviated as $M_o$. Diagonalizing $M_o$ yields, we obtain

$$U^{\dagger} M_o U = \Lambda , \tag{A11}$$

where the eigenmatrix $U = \frac{1}{\sqrt{3}}\begin{pmatrix} 1 & \varepsilon & \varepsilon^* \\ 1 & \varepsilon^* & \varepsilon \\ 1 & 1 & 1 \end{pmatrix}$, $U^{\dagger}$ stands for the conjugate transpose of the eigenmatrix $U$, and the eigenfrequencies $\Lambda = -\begin{pmatrix} 0 & & \\ & 3 & \\ & & 3 \end{pmatrix}$, $\varepsilon = e^{i2\pi/3}$. Thus, the node voltages can be expressed as $\boldsymbol{v} = v_0 \boldsymbol{\varphi}_0 + v_{S_1} \boldsymbol{\varphi}_{S_1} + v_{S_2} \boldsymbol{\varphi}_{S_2}$, where $\varphi_0 = (1,1,1)^T / \sqrt{3}$, $\varphi_{S_1} = (\varepsilon, \varepsilon^*, 1)^T / \sqrt{3}$, $\varphi_{S_2} = (\varepsilon^*, \varepsilon, 1)^T / \sqrt{3}$. They are the basis functions of the irreducible representations of the $C_3$ group. And $\varphi_0$ represents the constant basis functions of space, while $\varphi_{S_1} = (\varepsilon, \varepsilon^*, 1)^T / \sqrt{3}$ and $\varphi_{S_2} = (\varepsilon^*, \varepsilon, 1)^T / \sqrt{3}$, serving as the doubly degenerate states, can be chosen as the basis functions for the pseudospin space.

As shown in Fig. 1(f), we utilize an inductive circuit as an example to conduct a non-Abelian theoretical analysis. The Kirchhoff circuit equation of the circuit-example is given by:

$$\boldsymbol{I} = Y\boldsymbol{V} = \begin{pmatrix} Y_o & Y_{12} & 0 & 0 \\ Y_{12}^T & Y_o & Y_{23} & 0 \\ 0 & Y_{23}^T & Y_o & Y_{34} \\ 0 & 0 & Y_{34}^T & Y_o \end{pmatrix}\boldsymbol{V} , \tag{A12}$$

where $\boldsymbol{V} = (v_1, v_2, v_3, ..., v_{10}, v_{11}, v_{12})$, $\boldsymbol{I} = (i_1, i_2, i_3, ..., i_{10}, i_{11}, i_{12})$, $Y_0$ represents the spin module of $C_3$, $Y_{12}$, $Y_{23}$, $Y_{34}$ are the coupling connection modules that respectively link cell-*1* with cell-*2*, cell-*2* with cell-*3*, and cell-*3* with cell-*4*. Specifically,

$$Y_o |= y_{C_0} M_0 - (6 y_C + 1.5 y_R ) , \tag{A13}$$

$$Y_{12} \models y_{C_0} M_{\sigma_0} + y_{C_0} M_{\sigma_1}, \tag{A14}$$

$$Y_{23} \models y_{C_0} M_{\sigma_0} + y_{C_0} M_{\sigma_2}, \tag{A15}$$

$$Y_{34} \models y_{C_0} M_{\sigma_0} + y_{R_0} M_{\sigma_3}. \tag{A16}$$

Subsequently, by applying the *U* matrix to transform the above equation, we obtain:

$$\begin{pmatrix} \tilde{i}_{1,2,3} \\ \tilde{i}_{4,5,6} \\ \tilde{i}_{7,8,9} \\ \tilde{i}_{10,11,12} \end{pmatrix} = \begin{pmatrix} U^\dagger Y_0 U & U^\dagger Y_{12} U & 0 & 0 \\ U^\dagger Y_{12}^T U & U^\dagger Y_0 U & U^\dagger Y_{23} U & 0 \\ 0 & U^\dagger Y_{23}^T U & U^\dagger Y_0 U & U^\dagger Y_{34} U \\ 0 & 0 & U^\dagger Y_{34}^T U & U^\dagger Y_0 U \end{pmatrix} \begin{pmatrix} \tilde{v}_{1,2,3} \\ \tilde{v}_{4,5,6} \\ \tilde{v}_{7,8,9} \\ \tilde{v}_{10,11,12} \end{pmatrix}, \tag{A17}$$

where the eigenmatrix $U = \frac{1}{\sqrt{3}} \begin{pmatrix} 1 & \varepsilon & \varepsilon^* \\ 1 & \varepsilon^* & \varepsilon \\ 1 & 1 & 1 \end{pmatrix}$, $\varepsilon = e^{i2\pi/3}$, $U^\dagger$ stands for the conjugate transpose of the eigenmatrix *U*. Simultaneously, the current and voltage denoted with a tilde(~) at the top are calculated based on the characteristic functions of the $C_3$ symmetry group as the basis vectors, whereas the current and voltage without a tilde represent the actual current and voltage at the nodes of the circuit.

Here, we focus primarily on the right side of Eq. (A17), where the basis vectors have now transformed into $(\tilde{v}_{1,2,3}, \tilde{v}_{4,5,6}, \tilde{v}_{7,8,9}, \tilde{v}_{10,11,12})^T$, with $\tilde{\boldsymbol{v}} = U^\dagger \boldsymbol{v}$. Taking $U^\dagger Y_{12} U$ as an example, it is defined as:

$$\begin{aligned} Y_{T12} = U^\dagger Y_{12} U &= \begin{pmatrix} 2y_{C_0} & 0 & 0 \\ 0 & y_{C_0} & y_{C_0} \\ 0 & y_{C_0} & y_{C_0} \end{pmatrix} \\ &= (2y_{C_0}) \oplus \begin{pmatrix} y_{C_0} & y_{C_0} \\ y_{C_0} & y_{C_0} \end{pmatrix}, \end{aligned} \tag{A18}$$

where $y_{C_0}$ represents the admittance of $C_0$, $\oplus$ denotes the direct sum. The term preceding $\oplus$ is referred to as the constant representation space, and the subsequent term is termed the pseudo-spin space. Applying the same operation, we can derive $Y_{T0}$, $Y_{T23}$ and $Y_{T34}$. The operation transforms Eq. (A17) into a pseudo-spin space and reorders the bases, segregating the constant representation space from the pseudo-spin space, which is detailed below:

$$Y = \begin{pmatrix} \lambda_1 & t_1 & 0 & 0 \\ t_1 & \lambda_1 & t_2 & 0 \\ 0 & t_2 & \lambda_1 & t_3 \\ 0 & 0 & t_3 & \lambda_1 \end{pmatrix} \oplus \begin{pmatrix} \lambda_2 & h_{12} & 0 & 0 \\ h_{21} & \lambda_2 & h_{23} & 0 \\ 0 & h_{32} & \lambda_2 & h_{34} \\ 0 & 0 & h_{43} & \lambda_2 \end{pmatrix}, \tag{A19}$$

where

$$\lambda_{1S} |= -6sC_0 - 1.5/R_0$$
$$\lambda_{2S} |= \lambda_{1S} - 3sC_0$$
$$t_1 |= 2sC_0$$
$$t_2 |= 4sC_0$$
$$t_3 |= sC_0 + 3/R_0$$
$$h_{12} |= sC_0(\sigma_0 + \sigma_1)$$
$$h_{21} |= h_{12}$$
$$h_{23} |= s(C_0\sigma_0 + \sqrt{3}C_0\sigma_2)$$
$$h_{32} |= h_{23}$$
$$h_{34} |= s(C_0\sigma_0 + \sqrt{3}R_0 i\sigma_3)$$
$$h_{43} |= s(C_0\sigma_0 - \sqrt{3}R_0 i\sigma_3)$$

Continuing the derivation, we examine the relationship between the input current and the output voltage. It is known from $\boldsymbol{I} = \boldsymbol{YV}$ that

$$\boldsymbol{V} = \boldsymbol{ZI},\ Z = Y^{-1}, \tag{A20}$$

$$Z = Y^{-1} = \begin{pmatrix} \lambda_1 & t_1 & 0 & 0 \\ t_1 & \lambda_1 & t_2 & 0 \\ 0 & t_2 & \lambda_1 & t_3 \\ 0 & 0 & t_3 & \lambda_1 \end{pmatrix}^{-1} \oplus \begin{pmatrix} \lambda_2 & h_{12} & 0 & 0 \\ h_{21} & \lambda_2 & h_{23} & 0 \\ 0 & h_{32} & \lambda_2 & h_{34} \\ 0 & 0 & h_{43} & \lambda_2 \end{pmatrix}^{-1}. \tag{A21}$$

And we take the representation of space with constants as an example,

$$\begin{pmatrix} \lambda_1 & t_1 & 0 & 0 \\ t_1 & \lambda_1 & t_2 & 0 \\ 0 & t_2 & \lambda_1 & t_3 \\ 0 & 0 & t_3 & \lambda_1 \end{pmatrix}^{-1} = \begin{pmatrix} \frac{\lambda_1(-\lambda_1^2 + t_2^2 + t_3^2)}{\gamma} & \frac{t_1(\lambda_1^2 - t_3^2)}{\gamma} & \frac{-\lambda_1 t_1 t_2}{\gamma} & \frac{t_1 t_2 t_3}{\gamma} \\ \frac{t_1(\lambda_1^2 - t_3^2)}{\gamma} & \frac{-\lambda_1(\lambda_1^2 - t_3^2)}{\gamma} & \frac{\lambda_1^2 t_2}{\gamma} & \frac{-\lambda_1 t_2 t_3}{\gamma} \\ \frac{-\lambda_1 t_1 t_2}{\gamma} & \frac{\lambda_1^2 t_2}{\gamma} & \frac{-\lambda_1(\lambda_1^2 - t_1^2)}{\gamma} & \frac{t_3(\lambda_1^2 - t_1^2)}{\gamma} \\ \frac{t_1 t_2 t_3}{\gamma} & \frac{-\lambda_1 t_2 t_3}{\gamma} & \frac{t_3(\lambda_1^2 - t_1^2)}{\gamma} & \frac{\lambda_1(-\lambda_1^2 + t_1^2 + t_2^2)}{\gamma} \end{pmatrix}, \tag{A22}$$

where $\gamma = \lambda_1^2(-\lambda_1^2 + t_1^2 + t_2^2 + t_3^2) - t_1^2 t_3^2$. Subsequently, by substituting Eq. (A22) into the formula $\boldsymbol{V} = \boldsymbol{ZI}$, the calculation of the output voltage results in

$$(\tilde{v}_1, \tilde{v}_4, \tilde{v}_7, \tilde{v}_{10})^T = \begin{pmatrix} \frac{\lambda_1(-\lambda_1^2 + t_2^2 + t_3^2)}{\gamma} & \frac{t_1(\lambda_1^2 - t_3^2)}{\gamma} & \frac{-\lambda_1 t_1 t_2}{\gamma} & \frac{t_1 t_2 t_3}{\gamma} \\ \frac{t_1(\lambda_1^2 - t_3^2)}{\gamma} & \frac{-\lambda_1(\lambda_1^2 - t_3^2)}{\gamma} & \frac{\lambda_1^2 t_2}{\gamma} & \frac{-\lambda_1 t_2 t_3}{\gamma} \\ \frac{-\lambda_1 t_1 t_2}{\gamma} & \frac{\lambda_1^2 t_2}{\gamma} & \frac{-\lambda_1(\lambda_1^2 - t_1^2)}{\gamma} & \frac{t_3(\lambda_1^2 - t_1^2)}{\gamma} \\ \frac{t_1 t_2 t_3}{\gamma} & \frac{-\lambda_1 t_2 t_3}{\gamma} & \frac{t_3(\lambda_1^2 - t_1^2)}{\gamma} & \frac{\lambda_1(-\lambda_1^2 + t_1^2 + t_2^2)}{\gamma} \end{pmatrix} (\tilde{i}_1, \tilde{i}_4, \tilde{i}_7, \tilde{i}_{10})^T \tag{A23}$$

Here, we focus exclusively on the output voltage $\tilde{v}_{10}$, given that only input $\tilde{i}_1$ is active, denoted as $\tilde{i}_1 \neq 0$, $\tilde{i}_4 \neq 0$, $\tilde{i}_7 \neq 0$, $\tilde{i}_{10} \neq 0$. Therefore, $\tilde{v}_{10} = \frac{t_1 t_2 t_3}{\gamma} \tilde{i}_1$, corresponding to the constant representation space of $Z_{out}$ in Eq. (A21). By employing the same method, the relationship between $\tilde{v}_{11}$ and $\tilde{v}_{12}$ with respect to $\tilde{i}_2$ and $\tilde{i}_3$ in the pseudo-spin space can be determined. It is discovered that starting from Kirchhoff's equation $\boldsymbol{I} = \boldsymbol{Y}\boldsymbol{V} = \begin{pmatrix} Y_o & Y_{12} & 0 & 0 \\ Y_{12}^T & Y_o & Y_{23} & 0 \\ 0 & Y_{23}^T & Y_o & Y_{34} \\ 0 & 0 & Y_{34}^T & Y_o \end{pmatrix} \boldsymbol{V}$ and deriving based on $\tilde{i}_{4,5,6,\ldots,12} = 0$, identical results can be obtained as well. Upon completing the calculations, the resulting transfer function is as follows:

$$\tilde{v}_{out} = Z_{out}\tilde{i}_{in} = (t_1 t_2 t_3 / \gamma \oplus h_{21} h_{32} h_{43} / \Delta)\tilde{i}_{in}, \tag{A24}$$

where $\psi = \lambda_1^2(-\lambda_1^2 + t_1^2 + t_2^2 + t_3^2) - t_1^2 t_3^2$, $\Delta = \lambda_2^2(h_{12}h_{21} + h_{23}h_{32} + h_{34}h_{43} - \lambda_2^2) - h_{12}h_{21}h_{34}h_{43}$, $\tilde{v}_{out}$ represents the output voltage at cell-*4*, and $\tilde{i}_{in}$ denotes the input current at cell-*1*.

## APPENDIX C COMPREHENSIVE DIAGRAM OF ALL COUPLED CONNECTIONS

We have compiled a comprehensive list of all coupling connections. During data collection, it's ensured that all couplings in the $C_3$ circuit are selected from Fig. S2, and each type of coupling occurred no more than once.

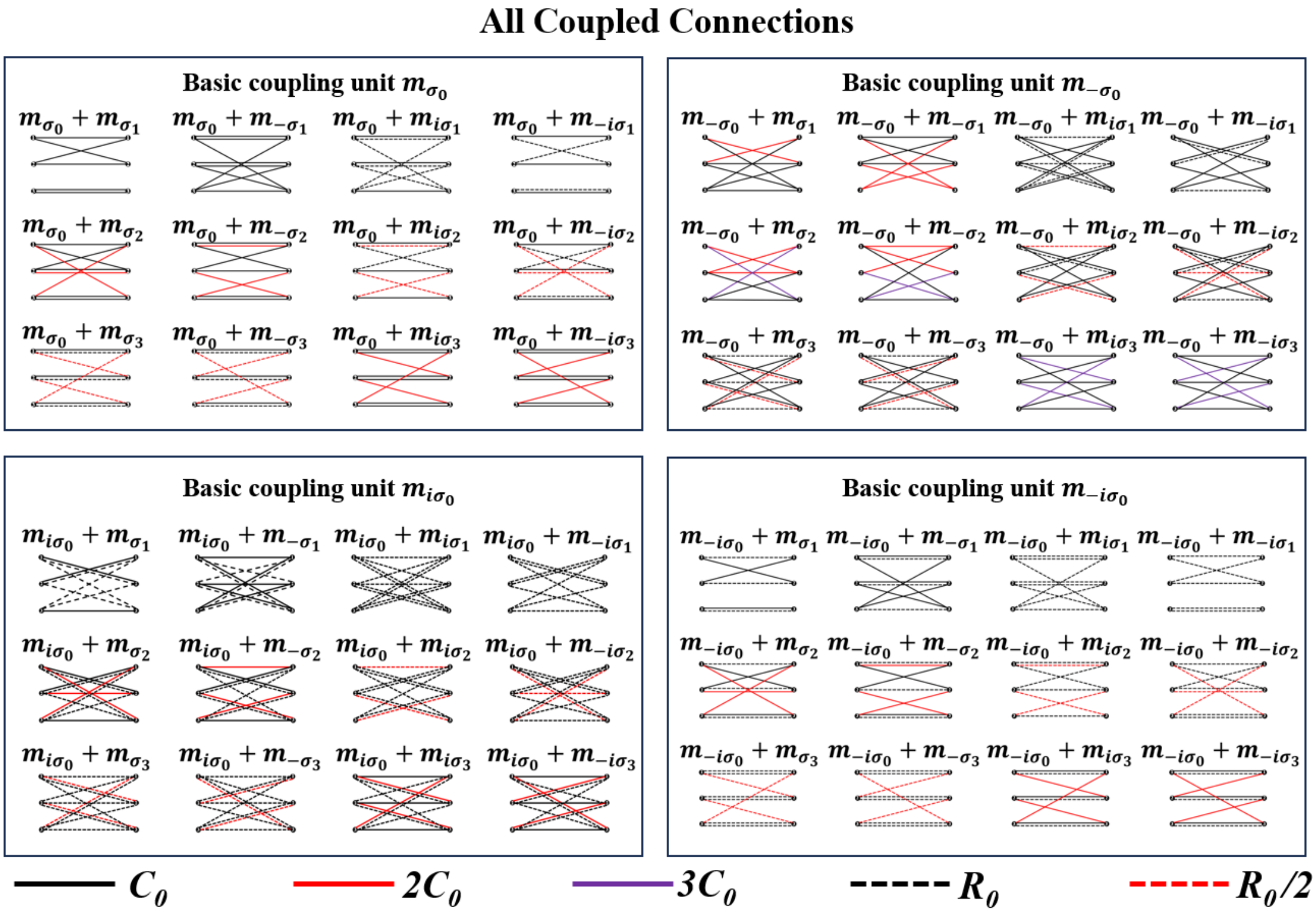

Fig. S2 Comprehensive diagram of all coupled connections.

We have illustrated the modular assembly of the tunneling matrices $m_{\pm(i)\sigma_{0,1,2,3}}$ in pseudospin space, where the solid lines indicate capacitors and the dashed lines represent resistors.

## APPENDIX D DETAILED CALCULATION PROCESSES OF THE TRANSFER FUNCTIONS FOR CIRCUITS 1, 2, AND 3

In Circuit 1, cell-*1* and cell-*2* are connected by modules $m_{\sigma_0}$ and $m_{\sigma_1}$. Within the spinor subspace, the relationship between cell-*1* and cell-*2* can be represented as $m_{01} = i\omega c(\sigma_0 + \sigma_1)$. The connection between cell-*2* and cell-*3* involves modules $m_{\sigma_0}$ and $m_{\sigma_2}$, with the mode of connection being $m_{02} = i\omega c(\sigma_0 + \sigma_2)$. The linkage between cell-*3* and cell-*4* is facilitated by modules $m_{\sigma_0}$ and $m_{\sigma_3}$, with their mode of connection defined as $m_{03} = i(\omega c\sigma_0 + R^{-1}\sigma_3) = i\alpha e^{\theta_3\sigma_3}$, where $\alpha = \sqrt{\omega^2 C^2 + 1/R^2}$, $\theta_3 = \arctan(1/\omega RC)$. Circuit-2 is derived by swapping the order of $m_{01}$ and $m_{02}$ in circuit-1, and circuit-3 is obtained by exchanging the positions of $m_{02}$ and $m_{03}$ in circuit-1. Given the non-commutativity between $m_{01}$ and $m_{02}$, as well as between $m_{02}$ and $m_{03}$, circuits 1, 2, and 3 are poised to produce unique outputs in response to an identical input signal. To rigorously validate this outcome, we consider the injection of current at cell-*1* and the detection of voltage at cell-*4*. The equations for the transfer impedance at the input and output of circuits 1, 2, and 3 are as follows:

$$\tilde{v}_{out} = (z_{1,2,3} \oplus Z_s^{1,2,3})\tilde{i}_{in} = (t_1 t_2 t_3 / \psi \oplus h_{21} h_{32} h_{43} / \Delta)\tilde{i}_{in}, \qquad \text{(A25)}$$

where $\tilde{v}_{out}$ represents the voltage measured at nodes 10, 11, and 12 of cell-*4*, $\tilde{i}_{in}$ denotes the current input at nodes 1, 2, and 3 of cell-*1*, $Z_{1,2,3}$ are constants denoting the spatial transfer impedance, $Z_s^{1,2,3}$ symbolizes the transfer impedance matrix within the pseudospin space, $\psi = \lambda_1^2(-\lambda_1^2 + t_1^2 + t_2^2 + t_3^2) - t_1^2 t_3^2$, and $\Delta = \lambda_2^2(h_{12}h_{21} + h_{23}h_{32} + h_{34}h_{43} - \lambda_2^2) - h_{12}h_{21}h_{34}h_{43}$.

For circuit-1,

$$\lambda_{1S} \models -6sC_0 - 1.5/R_0,\ \lambda_{2S} \models \lambda_{1S} - 3sC_0,\ t_1 \models 2sC_0,\ t_2 \models 4sC_0,\ t_3 \models sC_0 + 3/R_0,$$
$$h_{12} \models sC_0(\sigma_0 + \sigma_1),\ h_{21} \models h_{12},\ h_{23} \models s(C_0\sigma_0 + \sqrt{3}C_0\sigma_2),\ h_{32} \models h_{23},$$
$$h_{34} \models s(C_0\sigma_0 + \sqrt{3}R_0 i\sigma_3),\ h_{43} \models s(C_0\sigma_0 - \sqrt{3}R_0 i\sigma_3).$$

For circuit-2,

$$\lambda_{1S} \models -6sC_0 - 1.5/R_0,\ \lambda_{2S} \models \lambda_{1S} - 3sC_0,\ t_1 \models 4sC_0,\ t_2 \models 2sC_0,\ t_3 \models sC_0 + 3/R_0,$$
$$h_{12} \models s(C_0\sigma_0 + \sqrt{3}C_0\sigma_2),\ h_{21} \models h_{12},\ h_{23} \models sC_0(\sigma_0 + \sigma_1),\ h_{32} \models h_{23},$$
$$h_{34} \models s(C_0\sigma_0 + \sqrt{3}R_0 i\sigma_3),\ h_{43} \models s(C_0\sigma_0 - \sqrt{3}R_0 i\sigma_3).$$

For circuit-3,

$\lambda_{1S} \,|\!= -5sC_0 - 1.5/R_0$, $\lambda_{2S} \,|\!= \lambda_{1S} - 3sC_0$, $t_1 \,|\!= 2sC_0$, $t_2 \,|\!= sC_0 + 3/R_0$, $t_3 \,|\!= 4sC_0$,
$h_{12} \,|\!= sC_0(\sigma_0 + \sigma_1)$, $h_{21} \,|\!= h_{12}$, $h_{23} \,|\!= s(C_0\sigma_0 + \sqrt{3}R_0 i\sigma_3)$, $h_{32} \,|\!= s(C_0\sigma_0 - \sqrt{3}R_0 i\sigma_3)$,
$h_{34} \,|\!= s(C_0\sigma_0 + \sqrt{3}C_0\sigma_2)$, $h_{43} \,|\!= h_{34}$.

By substituting the parameters of circuits 1, 2, and 3 into Eq. (A21), respectively, the transfer function between the output voltage and the input current can be obtained.

## APPENDIX E EXPERIMENTAL MEASUREMENT PLATFORM FOR NON-ABELIAN NON-RECIPROCAL PROPERTIES IN REAL SPACE

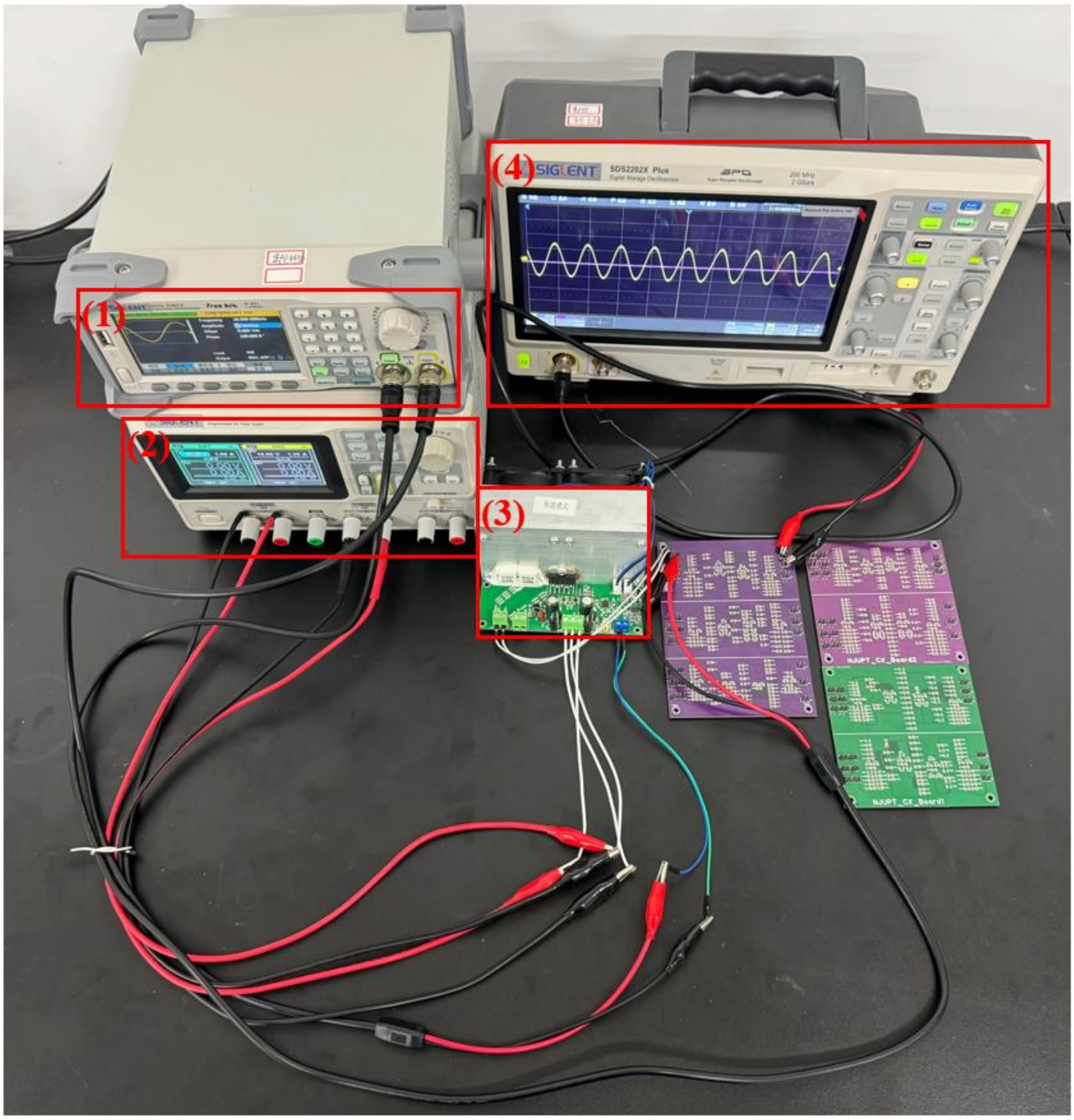

Fig. S3 (1) Programmable DC Power Supply (SIGLENT SPD3303X-E). (2) Function/Arbitrary Waveform Generator (SIGLENT SDG 2082X). (3) OPA549 Module for Transforming Motor. (4) Digital Storage Oscilloscope (SIGLENT SDS2202X Plus).